\documentclass[preprint,12pt]{elsarticle}
\makeatletter
\def\ps@pprintTitle{%
 \let\@oddhead\@empty
 \let\@evenhead\@empty
 \def\@oddfoot{}%
 \let\@evenfoot\@oddfoot}
\makeatother
\usepackage{graphicx}
\usepackage{amssymb}

\usepackage[utf8]{inputenc}
\usepackage{todonotes} 
\usepackage{amsmath}
\usepackage{hyperref}
\usepackage{float}
\usepackage{amsfonts}
\usepackage[margin=2.5cm]{geometry}
\usepackage{graphicx}
\usepackage{caption}
\usepackage{subcaption}
\usepackage{multirow}
\usepackage{makecell}
\usepackage{enumitem}

\begin{document}

\begin{frontmatter}

\title{Alignment and stability of embeddings: measurement and inference improvement}

\author[1,2]{Furkan Gürsoy\corref{cor1}}\ead{furkan.gursoy@boun.edu.tr}
\author[2,3]{Mounir Haddad}\ead{mounir.haddad@imt-atlantique.fr}
\author[2]{Cécile Bothorel}\ead{cecile.bothorel@imt-atlantique.fr}

\address[1]{Boğaziçi University, Department of Management Information Systems, Istanbul 34342, Turkey}
\address[2]{IMT Atlantique, LUSSI Department, Lab-STICC UMR CNRS 6285, Brest, France}
\address[3]{DSI Global Services, 41 avenue du Général Leclerc, Plessis-Robinson, France}

\cortext[cor1]{Corresponding author}

\begin{abstract}
Representation learning (RL) methods learn objects’ latent embeddings where information is preserved by distances. Since distances are invariant to certain linear transformations, one may obtain different embeddings while preserving the same information. In dynamic systems, a temporal difference in embeddings may be explained by the stability of the system or by the misalignment of embeddings due to arbitrary transformations. In the literature, embedding alignment has not been defined formally, explored theoretically, or analyzed empirically. Here, we explore the embedding alignment and its parts, provide the first formal definitions, propose novel metrics to measure alignment and stability, and show their suitability through synthetic experiments. Real-world experiments show that both static and dynamic RL methods are prone to produce misaligned embeddings and such misalignment worsens the performance of dynamic network inference tasks. By ensuring alignment, the prediction accuracy raises by up to 90\% in static and by 40\% in dynamic RL methods.
\end{abstract}


\end{frontmatter}


\section{Introduction}

Complex information about objects and relationships between them can be effectively and efficiently represented in low-dimensional latent spaces instead of their original representations in high-dimensional spaces. Examples of such complex systems include networks, text documents, or images. Representation learning methods are concerned with inferring such low-dimensional representations. These representations are also called embeddings. In the case of networks, one popular research question is to learn low-dimensional node embeddings from high-dimensional adjacency matrices. In natural language processing field, another popular research question is to learn word embeddings. In general, any pair of objects that are similar in the original system should also be closer in the latent space and vice versa. Therefore, embeddings preserve most of the useful information and can be ultimately utilized in various problems such as prediction of metadata and relations, or detection of clusters.

The reference system of a latent space, in general, has no physical meaning itself. The coordinates of a single object in the latent space, i.e., the numbers which make up the embedding vector of an object, have no standalone usefulness. The useful information comes from its relative position to the other objects. In simplified terms, the information is provided by the distances between nodes. It follows that transformations of a latent space that do not change the pairwise distances do not change the available information. Hence, in static systems such as static networks, one can obtain different embeddings that provide the exact same information, e.g., possibly based on initial conditions provided to a specific representation learning method. This phenomenon and issues associated with it do not arise for tasks on static systems since there is only a single latent space to represent all objects at once.

In contrast to static systems, dynamic systems contain also temporal information. For instance, over time, words might change their meaning or nodes might change their metadata. For the former, the distance between the embedding vectors of a word at two times might show its semantic drift. For the latter, the decision boundaries learned in a previous timestep for a classification task can be used to predict the node labels in a later timestep. However, to investigate such temporal evolution using embeddings, it is necessary to have a fixed reference system for the latent space over time. Once the latent space is fixed over time, it also becomes possible to find average embeddings over time or to visualize temporal changes in the latent space.

For a given dynamic system, the difference between the learned representations for different timesteps can be caused either by actual changes in the system or by the transformations that do not alter the pairwise distances in the latent space. The first is related to \textit{stability} of the system over time whereas the second is related to the \textit{alignment} of embeddings which is an artifact of the utilized representation learning method and details of its implementation. For instance, a slightly changing system might produce drastically different embeddings at each timestep when embeddings at each timestep are learned separately by a static representation learning method because such small changes might significantly change the regions in which the learned embeddings converge. Moreover, even a system that does not change over time might result in different embeddings in different timesteps due to the stochasticity in many representation learning methods. In summary, stability characterizes the dynamics of the system itself, whereas misalignment characterizes spurious changes in the latent space. Therefore, dynamic embeddings are said to be \textit{aligned} when the difference between different timestep embeddings reflect the actual changes in the system rather than the ineffectual transformations of the latent space.

Overall, the issue of stability and alignment of dynamic embeddings carries utmost importance in tasks that make use of the temporal changes in the latent space such as future metadata prediction, temporal evolution, dynamic visualization, obtaining average embeddings, etc. Although acknowledged in the literature, to the best of our knowledge, the issue has not been defined formally, explored theoretically, or analyzed empirically in a sufficient manner.

In this study, we make the following contributions:

\begin{itemize}
  \item We dismantle the alignment and explore and explain its parts; and provide the first formal definition for embedding alignment.
  
  \item We develop appropriate and mathematically-justified metrics to measure the alignment and stability of dynamic embeddings and show their usefulness through a series of extensive synthetic and real-world experiments.
  
  \item We employ more than 10 various static and dynamic network representation learning methods and analyze alignment and stability metrics as well as inference task performance on seven real-world datasets with different characteristics.
  
  \item We show that when embeddings are forced to be aligned based on provided formal definitions, remarkable improvements in accuracy are obtained in dynamic network inference tasks.
  
\end{itemize}

The paper is structured as follows. In the next section, we explore the topic of alignment in more detail, look at its description in the literature, and review the existing approaches. In Section 3, we provide formal definitions for embedding stability and embedding alignment and propose a set of metrics to measure them. In Section 4, we conduct extensive synthetic experiments to show the appropriateness and behavior of the proposed measures. In Section 5, we conduct experiments using several real-world datasets and many embedding methods to test our approach in real-world cases and explore the relationship between alignment, stability, and dynamic inference task performance. The conclusion and final remarks are given in Section 6.

In the rest of the study, particularly in the real-world experiments, we focus on the case of dynamic network node embeddings. However, our proposed measures require only the embedding matrices and are agnostic to how and from what type of system these embeddings are created. Therefore, the proposed measures are directly generalizable to other types of embeddings such as subgraph embeddings, entire graph embeddings, word embeddings, document embeddings, image embeddings, and so on. Even more, our measures may be used outside the dynamic system assumption, e.g., to obtain average embedding vectors from different representation learning methods in a static system.

\section{Background}

First inspired by the advances in the natural language processing methods, graph representation learning techniques have seen a surge in the last years regarding their performances and the possibilities they offer. Graph embedding techniques can be classified into different categories. Some of them are based on node similarity matrix factorization \cite{cao2015grarep,ou2016asymmetric,belkin2002laplacian}. Another type adapts the notion of random walks to graph data to form sequences of nodes, similar to sentences within word embedding techniques \cite{grover2016node2vec,perozzi2014deepwalk}. Also, some methods take advantage of the advances in deep learning \cite{cao2016deep,kipf2016semi,wang2016structural}.

\begin{figure}[!h]
  \centering
  \begin{subfigure}[b]{0.275\textwidth}
   \includegraphics[width=\textwidth]{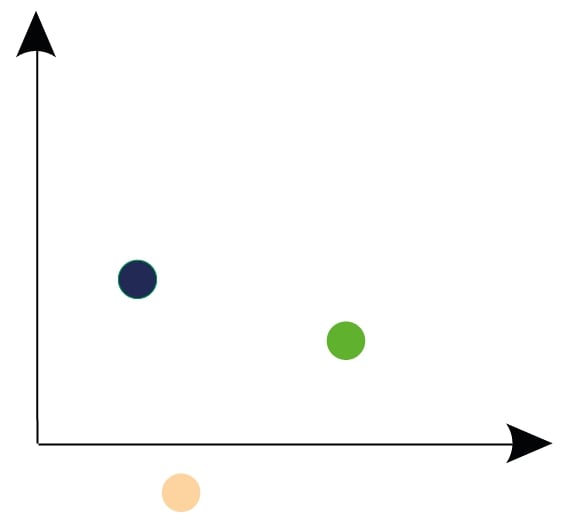}
   \caption{}
  \end{subfigure}
  \hspace{35mm}
  \begin{subfigure}[b]{0.275\textwidth}
   \includegraphics[width=\textwidth]{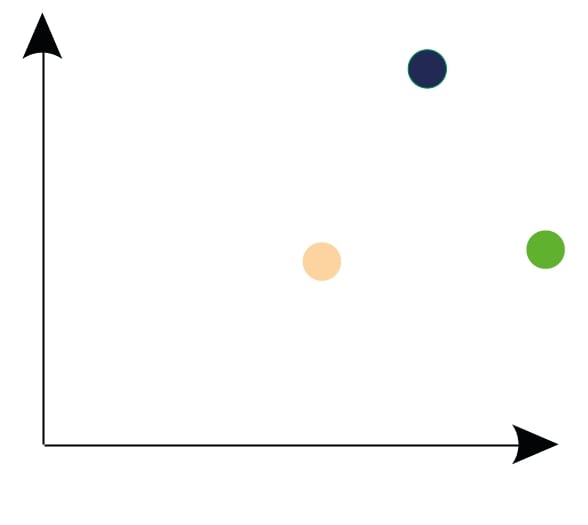}
   \caption{}
  \end{subfigure}
  \caption{An example of misalignment. (a) Sample embeddings at time $t$. (b) The embeddings for the same objects at time $t+1$. Although pairwise distances are preserved between the two timesteps, the embedding vectors are different. It demonstrates a misalignment where the difference is fully induced by a translation and a rotation. }
  \label{misalignment}
\end{figure}

Although those methods were ground-breaking, they are static as they were conceived to address graphs that do not evolve over time. When a static method is applied to compute dynamic network representations over its timesteps, the produced embeddings do not abide by any temporal coherence. As a matter of fact, pairwise euclidean distances between node embeddings remain unchanged regarding some operations like translation, rotation, or reflection. This means that, as described in Figure \ref{misalignment}, two consecutive timesteps static methods embeddings can be shifted or rotated one regarding the other as they might be placed in different latent spaces. In such situation, temporal embeddings are said to be misaligned.

Few dynamic approaches handle time (and alignment consequently) in the conception of the embeddings. Some approaches apply an optimal linear transformation minimizing the distance between the consecutive independently learned timesteps embeddings \cite{kulkarni2015statistically,zhou2019}. Some others initialize the current timestep representation with the previous embedding vectors within the learning process \cite{mahdavi2018dynnode2vec}. 
It is also possible to force embeddings continuity over time \cite{haddad2019temporalnode2vec}.

The main advantage of the dynamic embedding methods lies in the incorporation of temporal information to conceive more reliable global embeddings. However, alignment is an important issue that most of these methods do not generally address in particular. Some literature articles mention the alignment issue without providing an explicit way to handle it \cite{liang2018,palmucci2019your}. In word embeddings, \cite{kulkarni2015statistically} uses locally linear regression to align the neighborhood of a focal word in different timesteps. However, this operation has to be applied for each focal word independently. Some studies \cite{xu2018, singer2019node} employ orthogonal procrustes analysis: the idea is to find and apply the orthogonal matrix (i.e. rotation and/or reflection) that maps the two consecutive timesteps embeddings most closest to each other using singular value decomposition \cite{schonemann1966generalized}. Others \cite{fang2011graph, passino2019link, Hewapathirana2020} employ generalized procrustes analysis where optimal translation and scaling operations are allowed in addition to the rotation \cite{Gower1975}. It should be noted that orthogonal procrustes analysis does not offer the possibility of finding possible translations between timesteps embeddings, which might be a potential drawback when the distance metric to be considered in the embedding latent space is the euclidean distance rather than the dot product or the cosine similarity.

We consider two sets of embeddings as being aligned if there is no global move (shift, rotation, reflection) between the vectors they are composed of. In addition to its practical aspect, this definition flows from an observation one can make: in all its possible forms (graphs, adjacency matrices, edge lists...), the input data of the common embedding methods represent interactions between nodes; consequently, in terms of nodes embeddings dynamics, if the alignment issue is intuitively seen as movements of nodes, then a global move of embeddings should not be allowed as the input data cannot include such information. Based on that, we also define alignment forcing and measuring as the process of finding and quantifying the optimal linear transformations that bring the two timesteps embeddings to the closest.

Few works in the literature address the problem of measuring embedding alignment. \cite{Chen2019DynamicNE,goyal2018dyngem} define a quantity named embeddings temporal stability (or continuity) that is meant to measure the amount of \textit{misalignment} between two timesteps embeddings. The general idea is to analyse the correlation between the change of a node's neighborhood and the move over time of a node in the latent space. However, this quantity encompasses what we consider to be two related but different notions: alignment and stability. Stability reflects the changes that may occur between consecutive timesteps embeddings that are not attributable to alignment issues. For an ideally aligned embedding method, stability is directly related to the dynamics of the considered network,i.e., its structural evolution at different scales. It should also be noted that instability may also be caused by the used embedding method due to the randomness involved in the process of representation creation.

\section{Stability and Alignment}

In this section, we introduce and define the proposed alignment measures, as well as the strategy used to force alignment between different embeddings. As mentioned before, multiple operations that may occur between consecutive timesteps embeddings can compromise the alignment. So is the case with translations, rotations, and reflections. Then, an appropriate measure of the alignment should quantify the amplitude of each one of these operations. However, trying to measure alignment with a single value may hide different situations: for example, a rotation of some angle $\theta$ and a translation of vector $\Vec{t}$ may produce the same alignment measure. Such equivalencies do not seem to be relevant. Instead, we choose to design an alignment measure for each of the operations that come into play in alignment.

\subsection{Translation error}
This measure intends to quantify the average global shift between two consecutive timesteps embeddings. More specifically, we are interested in the move of the center of gravity across time. Given two consecutive embeddings $\{e^t_i\}$ and $\{e^{t+1}_i\}$ for a node $i \in [\![ 1, \:|V|]\!]$, we define the global shift as being:

\begin{equation}
\label{tglob}
\newcommand\norm[1]{\left\lVert#1\right\rVert}
t_{glob} = \norm{o^{t+1} - o^{t}} \: \: \: \: \: where \: \: \: \: \: o^{t} = \frac{\sum_{i=1}^{|V|} e^{t}_i}{|V|}
\end{equation}

As this value is not representative due to its dependence on the embedding space, one has to normalize it regarding some characteristic distances of the embeddings in the latent space. In our case, we choose the radius,i.e., the average distances to the center of gravity: 

\begin{equation}
\label{tnorm}
\newcommand\norm[1]{\left\lVert#1\right\rVert}
t_{norm} = \frac{t_{glob}}{r^t + r^{t+1}} \: \: \: \: \: where \: \: \: \: \: r^t = \frac{\sum_{i=1}^{|V|} \norm{e^{t}_i - o^{t}}}{|V|} 
\end{equation}

\begin{figure*}[]
\begin{center}
\includegraphics[scale=.25]{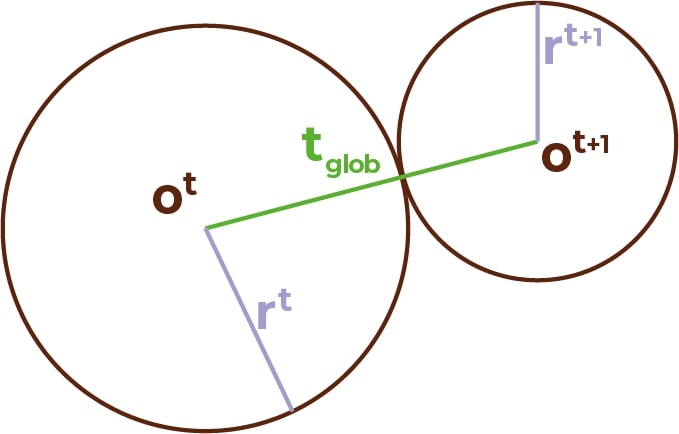}
\caption{Tangent embeddings regarding their radii. $o^t$, $r^t$ and $t_{glob}$ respectively refer to the center of gravity of the embeddings of timestep $t$ (defined in Equation \ref{tglob}), their radius (Equation \ref{tnorm}) and the global shift (Equation \ref{tglob}). In such case, $t_{norm}$ is equal to 1 and the translation error $\xi_{tr}$ is equal to 0.5.}
\label{tangentEmbeddings}
\end{center}
\end{figure*}

It is worth noting that, as Figure \ref{tangentEmbeddings} shows, $t_{norm}$ is equal to 1 when both timesteps embeddings are \textit{tangent} in terms of their radii.

The quantity $t_{norm}$ is limitless as it can reach infinite values. To bound our final translation error, we define: 

\begin{equation}
\xi_{tr} = \frac{t_{norm}}{t_{norm} + 1}
\end{equation}

Therefore, $\xi_{tr}$ ranges between 0 and 1. We also mention that the value of the $\xi_{tr}$ quantity depends on the involved embeddings. This means that comparing shift errors for different temporal network embeddings can be misleading.

Finally, once the translation error is measured, we center the two compared embeddings by shifting them to make their centers of gravity match with the origin of the embedding latent space: 

\begin{equation}
C^t = \{c^t_i\} \: \: \: \: \: with \: \: \: \: \: c^{t}_i = e^{t}_i - o^{t} \: \: \: \: \: for \: \: \: \: \: i \in [\![ 1, \:|V|]\!] .
\end{equation}

\subsection{Rotation error}
Given two centered consecutive timesteps embeddings $C^t$ and $C^{t+1}$, it is possible to find the optimal global rotation and/or reflection matrix $R$ that maps most closely $C^t$ with $C^{t+1}$. This can be achieved using orthogonal procrustes: 

\begin{equation}
\newcommand\norm[1]{\left\lVert#1\right\rVert}
R = argmin_{\Omega} \norm{C^t \: \Omega - C^{t+1}}
\end{equation}

To quantify the rotation/reflection between the embeddings, we compare $R$ to the identity matrix $I$: in the ideal case where $R = I$, $C^t$ and $C^{t+1}$ are already well oriented one regarding the other. We define the global rotation measure $\xi_{rot}$ as follows: 

\begin{equation}
\newcommand\norm[1]{\left\lVert#1\right\rVert}
\xi_{rot} = \frac{\norm{R - I}_F}{2\sqrt{d}} = \frac{\sqrt{2\:d - 2\:Tr(R)}}{2\sqrt{d}}
\end{equation}
where $d$ is the embedding latent space dimension and $Tr(R)$ is the trace of the matrix $R$. Dividing by $2\sqrt{d}$ brings $\xi_{rot}$ range from $0$ to $1$. That being set, one might well ask the question of the ability of $\xi_{rot}$ to quantify the \textit{amount} of rotation/reflection that $R$ represents. In order to answer this question, we introduce $R_{can}$, the canonical form of $R$: each orthogonal matrix can be decomposed into $a$ elementary rotations and brought into a block diagonal matrix form, (e.g., using the Schur decomposition):

\begin{equation}
R_{can} = Q^T \: R \: Q = 
\begin{bmatrix}
  R_1 & & & & & \\
  & \ddots & & & & \\
  & & R_a & & & \\
  & & & \pm 1 & & \\
  & & & & \ddots & \\
  & & & & & \pm 1
 \end{bmatrix}
\: \: \: \: \: with \: \: \: \: \: R_j = 
\begin{bmatrix}
cos(\theta_j) & -sin(\theta_j) \\
sin(\theta_j) & cos(\theta_j)
\end{bmatrix}
\end{equation}

and $Q$ is an orthogonal matrix. More precisely, as Table \ref{tab:canonicalForms} shows, the form of $R_{can}$ depends on the parity of $d$ and the sign of the determinant of $R$ (i.e. whether $R$ is a rotation only or it is a reflection/rotation):

\begin{table}[!h]
\footnotesize
\centering
\setlength{\tabcolsep}{1em} 
\renewcommand{\arraystretch}{1}
\caption{Orthogonal matrix canonical form}
\begin{tabular}{| c || c | c |}
\hline
 & \textbf{odd dimension} & \textbf{even dimension} \\
\hline
\makecell{\textbf{rotation only} \\ \textbf{det(R) = 1}} & $
\begin{bmatrix}
  R_1 & & & \\
  & \ddots & & \\
  & & R_{\frac{d-1}{2}} & \\
  & & & 1
 \end{bmatrix}
$ & $
\begin{bmatrix}
  R_1 & & \\
  & \ddots & \\
  & & R_{\frac{d}{2}}
 \end{bmatrix}
$ \\
\hline
\makecell{\textbf{rotation/reflection} \\ \textbf{det(R) = -1}} & $
\begin{bmatrix}
  R_1 & & & \\
  & \ddots & & \\
  & & R_{\frac{d-1}{2}} & \\
  & & & -1
 \end{bmatrix}
$ & $
\begin{bmatrix}
  R_1 & & & & \\
  & \ddots & & & \\
  & & R_{\frac{d-2}{2}} & & \\
  & & & 1 & \\
  & & & & -1
 \end{bmatrix}
$ \\
\hline
\end{tabular}
\label{tab:canonicalForms}
\end{table}

It is possible to look at $R_{can}$ as an expression of $R$ in a basis where it can be decomposed into $a$ elementary 2-dimensional rotations in orthogonal planes (with $\{\theta_j, j \in [\![ 1, \:a]\!]\}$ as rotation angles), and possibly a reflection. Both matrices $R$ and $R_{can}$ have the same rotation error $\xi_{rot}$ as they share the same trace. Also, as the trace of $R_{can}$ is related the $\theta_j$ angles, $\xi_{rot}$ can be written as following: 

\begin{equation}
\xi_{rot} = \frac{\sqrt{2d - 4\sum_{j=1}^{A} cos(\theta_j) - 2\mu }}{2\sqrt{d}}
 \: \: \: \: \: where \: \: \: \: \: 
 \mu = \left\{
  \begin{array}{ll}
    1 & \mbox{if $d$ is odd and $\det(R) = 1$} \\
    -1 & \mbox{if $d$ is odd and $\det(R) = -1$} \\
    0 & \mbox{if $d$ is even }
  \end{array}
\right.
\end{equation}

Thus, we can observe that the smaller $\theta_j$ angles are, the smaller $\xi_{rot}$ is. Also, reflections increase the rotation error.

Lastly, after measuring the rotation/reflection error, we rotate $C^t$ embedding matrix accordingly to $R$ matrix in order to obtain centered well oriented consecutive timesteps embeddings: 

\begin{equation}
C^t_r = C^t \: R
\end{equation}

We consider the centered well-oriented matrices $C^t_r$ and $C^{t+1}$ to be the aligned versions of the input embeddings.

\subsection{Scale error}
This measure intends to examine the change of scale over time of centered well-oriented embeddings. For this purpose, we define the scale error based on the radii,i.e., the average distances to the origin: 

\begin{equation}
  \xi_{sc} = \frac{|r^t - r^{t+1}|}{r^t + r^{t+1}}
\end{equation}

It ranges from 0 to 1. In particular, $\xi_{sc}$ is nil when the embeddings have the exact same radius. To the contrary, $\xi_{sc}$ approaches 1 when the radii of the embeddings are very different. Once the scale error is measured, we normalize the centered well oriented embeddings by their respective radii: 

\begin{equation}
  N^t = \frac{C^t_r}{r^t}
\end{equation}

It should be noted that we do not consider the scale error as being part of the alignment. As a matter of fact, contrary to translation, rotation, and reflection, pairwise distance matrices are not invariant under the change of scale. However, the change of scale can indicate the evolution of the density of embeddings and, consequently, the change of the strength of the interactions between the input graph nodes.

\subsection{Stability error} 
Given two centered, well oriented and normalized embeddings, it is interesting to examine the evolution of embeddings over time. Indeed, the difference between the respective characteristics of $N^t = \{n^t_i\}$ and $N^{t+1} = \{n^{t+1}_i\}$ cannot be attributable to alignment issues. Thus, the comparison between $N^t$ and $N^{t+1}$ gives information about the structural dynamics of the embedded network, like the temporal continuity of the embeddings or the noise involved between consecutive timesteps. We define the stability error as follows: 

\begin{equation}
\newcommand\norm[1]{\left\lVert#1\right\rVert}
\xi_{st} = \frac{ \sum_{i=1}^{|V|} \frac{\norm{n^{t+1}_i - n^{t}_i}}{ \norm{n^{t}_i} + \norm{n^{t+1}_i} } }{|V|}
\end{equation}

The stability error $\xi_{st}$ ranges from 0 to 1. It is equal to 0 if and only if $N^t = N^{t+1}$.

Even though the stability error is computed after the alignment and scaling measuring process, it brings additional information about the relevance of the other defined measures. For example, in the extreme case of comparing two randomly generated embedding matrices, one can still obtain $\xi_{tr}$, $\xi_{rot}$ and $\xi_{sc}$. However, those values are meaningless. In that case, the stability error should be relatively high.

\section{Experiments on synthetic networks}
\label{syntheticNetworks}

We design and conduct extensive synthetic experiments to demonstrate the suitability and characteristics of the proposed error measures. The behavior of each error measure is explored under interesting combinations of transformations such as scaling, translations, rotations, and noise; by applying them to an initial matrix (e.g., $E^t$) to create the transformed matrix (e.g., $E^{t+1}$). The error measures are then computed to investigate the stability and alignment of the two embeddings, where we can trace the differences directly to the applied synthetic transformations. We are also interested in whether the proposed measures are affected by the shape of the embedding matrix, represented by the number of objects and the length of embedding vectors.

\begin{figure}[!h]
  \centering
  \begin{subfigure}[b]{0.47\textwidth}
   \includegraphics[width=\textwidth]{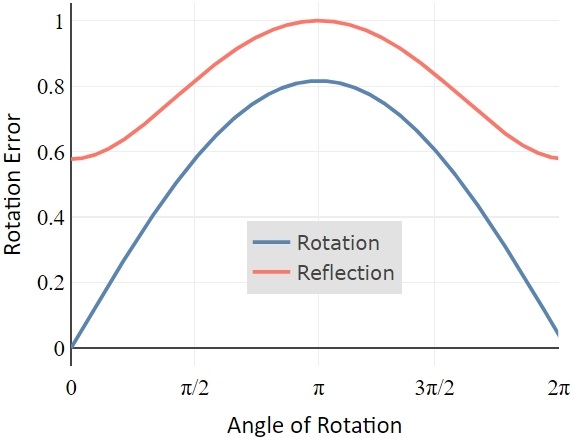}
   \caption{}
   \label{subfig:rot3d}
  \end{subfigure}
  \hspace{5mm}
  \begin{subfigure}[b]{0.47\textwidth}
   \includegraphics[width=\textwidth]{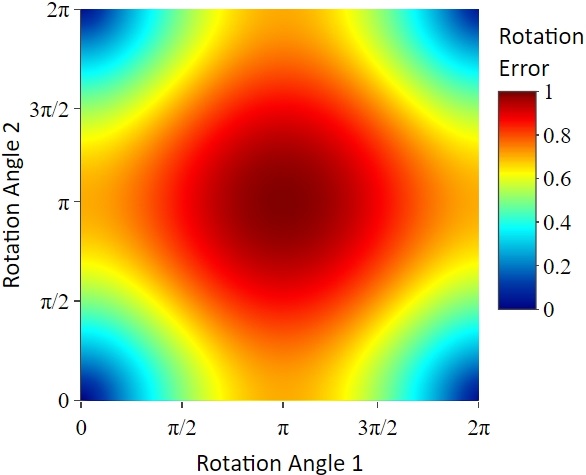}
   \caption{}
   \label{subfig:rot4d}
  \end{subfigure}
  \caption{Magnitude of rotation versus the rotation error. (a) The angle of rotation, an optional reflection, and corresponding rotation error when $d=3$. (b) The two angles of rotation and corresponding rotation error when $d=4$. Overall, rotation error reflects magnitudes of rotation angles and the optional reflection.}
  \label{fig:rotationangles}
\end{figure}

Any rotation matrix for an embedding matrix can be represented in its canonical form with $d/2$ angles or $(d-1)/2$ angles and an optional reflection when $d$ is even, or with $(d-1)/2$ angles and a reflection when $d$ is odd. Intuitively, we expect our rotation error to reflect the magnitude of the optimal rotation matrix as measured by its underlying angles. To test this, and to provide a visual demonstration for the cases when $d=3$ and $d=4$, we create a large number of rotation matrices with varying rotation angles. Figure \ref{subfig:rot3d} visualizes the obtained rotation error for different angles using two separate series for the cases where reflection is applied or not. The rotation error is at its maximum when the rotation angle is maximized, i.e., when it is equal to $\pi$ regardless of whether the reflection is applied or not. We also observe that the application of reflection on top of any angle of rotation worsens the error. As depicted in Figure \ref{subfig:rot4d} the rotation error is maximized when both angles are equal to $\pi$, and lowest when both angles are $0$. Overall, Figure \ref{fig:rotationangles} shows the clear relationship between the angles of rotation and reflection and the proposed rotation error, visually confirming the intuitive expectations and the mathematical implications.

\begin{figure}[!h]
  \centering
  \begin{subfigure}[b]{0.42\textwidth}
   \includegraphics[width=\textwidth]{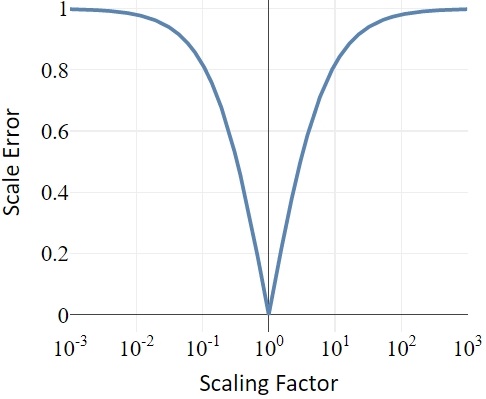}
   \caption{}
   \label{subfig:scalefactorerror}
  \end{subfigure}
  \hspace{0.06\textwidth}
  \begin{subfigure}[b]{0.42\textwidth}
   \includegraphics[width=\textwidth]{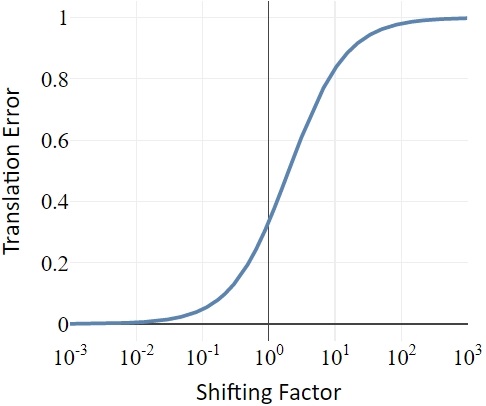}
   \caption{}
   \label{subfig:shiftfactorerror}
  \end{subfigure}
  \caption{Scale and translation errors versus magnitudes of respective transformations. (a) Scaling factor and corresponding scale error. (b) Shifting factor and corresponding translation error. Overall, scale and translation errors reflect the magnitudes of respective transformations.}
  \label{fig:scaleshift_factorerror}
\end{figure}

When two embeddings are the same except that one is a scaled-up (i.e., dilated) or a scaled-down (i.e., contracted) version of the other, we expect the scaling error to reflect the magnitude of scaling. To demonstrate this, we create a large number of initial embedding matrices, apply scale transformations with different magnitudes of scaling factors, and report the results. To create an initial embedding matrix, the number of objects $n$ is uniformly sampled from the range $[10, 10000]$, the length of the embedding vectors $d$ is uniformly sampled from the range $[2, 32]$, and the matrix is filled with $n \times d$values uniformly sampled from the range $[0, 1]$. Each initial embedding matrix is then scaled with different scaling factors such that the center of gravity is preserved and the ratio between radiuses of the transformed and initial embedding matrices is equal to the scaling factor. Figure \ref{subfig:scalefactorerror} reports the results of this experiment; and demonstrates that the scale error is $0$ when two embeddings have the same scale (i.e., when the scaling factor is $1$ hence the radiuses are equal) and approaches $1$ as the scaling factor moves away from $1$ in either direction (e.g., dilation or contraction). 

In a similar fashion, when two embeddings are the same except that one is shifted in some direction in the space, we expect the translation error to reflect the magnitude of the shift. To demonstrate this, again, we create a large number of initial embedding matrices with the same sampling procedure. These matrices are then shifted in random directions such that the ratio of the radius of the initial matrix to the Euclidean norm of the shift vector is equal to the shifting factor. Figure \ref{subfig:shiftfactorerror} reports the result of this experiment; and demonstrates that the translation error is $0$ when the shifting factor is $0$ (i.e., the case of no translation) and approaches $1$ as the shifting factor increases. Overall, Figure \ref{fig:scaleshift_factorerror} shows graphically that the bounds of scale and translation errors are meaningful, the values within their range directly reflect the magnitudes of respective transformations as they should, and the measures are independent of the shape of the embeddings; providing the necessary properties we expect from such measures.

\begin{figure}[!h]
  \centering
  \begin{subfigure}[b]{0.48\textwidth}
   \includegraphics[width=\textwidth]{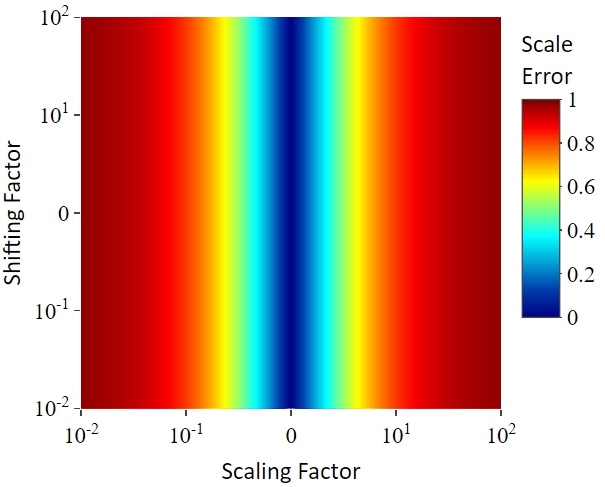}
   \caption{}
   \label{subfig:scale_err}
  \end{subfigure}
  \hfill
  \begin{subfigure}[b]{0.48\textwidth}
   \includegraphics[width=\textwidth]{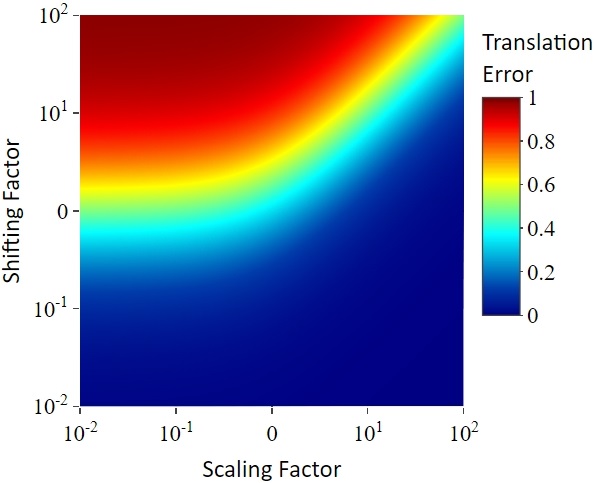}
   \caption{}
   \label{subfig:trans_err}
  \end{subfigure}
  \caption{Combined scaling and shifting versus the scale and translation errors. (a) Scaling and shifting factors, and corresponding scale errors. (b) Scaling and shifting factors, and corresponding translation errors. Overall, effects of combined scaling and shifting operations with different magnitudes on the scale and translation errors show that scale error reflects only the magnitude of scaling whereas translation error reflects the scale in addition to the magnitude of shifting due to the normalization by average radius.}
  \label{fig:scale_shift_combo}
\end{figure}

Next, we investigate the behavior of our error measures when the difference between the two embeddings can be explained only by scaling and shifting together. The initial embeddings are created and shifting and scaling are applied, all following the procedure employed for the previous experiment. In the case of scale error, we expect it to be independent of shifting since it is concerned with their radius of the embeddings and not where the embeddings are in the space. Confirming this, Figure \ref{subfig:scale_err} depicts that the scale error increases if and only when the scaling factor moves away from $1$. However, the translation error does not exclusively depend only on how distant two embeddings are in the space since such distance is meaningful and comparable with respect to the radiuses of the embeddings in question. Figure \ref{subfig:trans_err} visualizes the experiment where initial embedding matrices shifted first and scaled later with the respective factors. Confirming the expected behavior, the figure shows that the translation error is near $0$ when the shifting factor is near $0$. It increases as the shifting factor increases; and decreases at the same time when a large scaling factor increases the radius of the transformed matrix. In general, the smoothness of (i.e., lack of randomness in) patterns observed in Figure \ref{fig:scale_shift_combo} visually supports the earlier finding that two measures are independent of the shape of embeddings, hence generalizable across systems with different numbers of objects and different embedding vector lengths.

The experimental results along with the mathematical definitions provided earlier show that the effects of the three linear transformations are captured by the error measures specifically designed for them and the resulting error values are directly related to the magnitude of respective transformations. When the only difference between two embeddings is a rotation, the radius and the center of gravity do not change; hence only the rotation error produces a nonzero value. Likewise, if the only change is the radius or the center of gravity, only the respective scale or translation error produces a nonzero value. Accordingly, we can conclude that each of these measures only monitors the transformations relevant to them. The same is true when the difference between embeddings can be explained by a combination of these transformations since these well-separated transformations do not change the aspects of data monitored by the other error measures. The only exception to that is the translation error since it employs the average of the radiuses as its unit to be comparable across different settings. In particular, the value in its numerator still stays the same under non-shift transformations but its denominator changes with scaling. In general, these properties allow us to treat different error measures independently from each other. Specifically, we can reduce a selected nonzero error to zero by applying the respective reverse transformation and other error measures will not behave unexpectedly, i.e., their values will be exactly the same except for the translation error which will change in a predictable manner if the applied reverse transformation is scaling.

Unlike the previous three measures, the stability error evaluates and captures the changes in structure, more specifically the relative structural changes. It corrects for the rotation, translation, and scaling since the relative structure does not change under these transformations. A difference that can be explained fully by any combination of these transformations cannot produce a nonzero stability error. To investigate the behavior of stability with synthetic experiments, we introduce structural changes to the embeddings by adding noise. Specifically, the addition of noise will change the distances between objects in the latent space. 

The noise addition procedure is modeled as a random walk in $d$-dimensional space where the number of random walk steps corresponds to the noise factor. At each random walk step, the embedding matrix $E^{n \times d}$ is updated following Equation \ref{eq:noise} where $U(a,b)^{n \times d}$ is a noise matrix with the same shape as embeddings whose values uniformly sampled from the range $(a,b)$. This ensures that all steps are random across directions and across objects. The term $\sqrt{d}$ ensures that the noise factor is equivalent across different embedding shapes. $r/2$ is a practical choice that ensures that step size is neither too small nor too large. It follows that $25$ random walk steps result in added noise with approximately the same radius as the original embedding since $\mathbb{P}(\sum_{1 to 25} \frac{u(-1,1)}{2} > 1) \simeq \mathbb{P}(\sum_{1 to 25} \frac{u(-1,1)}{2} < 1) \simeq 50\%$ where $u(-1, 1)$ is a scalar uniformly sampled from the respective range. Accordingly, we set $25$ steps as the unit for the noise factor, i.e., noise factor of $1$ corresponds to noise with the same radius as the original embedding on average.

\begin{equation}
\label{eq:noise}
E \leftarrow E + \frac{U(-1,1) r}{2 \sqrt{d}}
\end{equation}

\begin{figure}[!h]
  \centering
  \begin{subfigure}[b]{0.42\textwidth}
   \includegraphics[width=\textwidth]{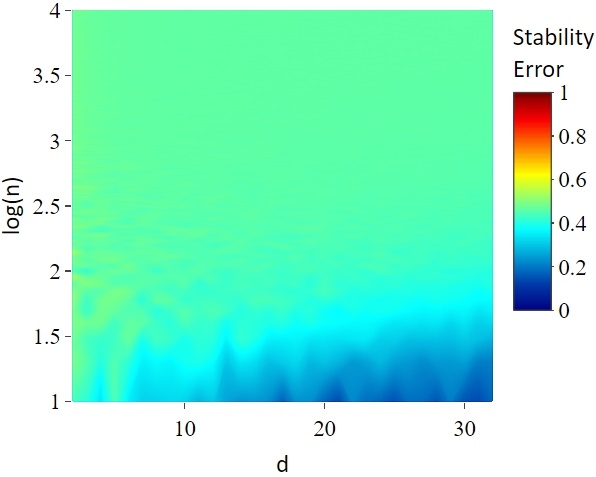}
   \caption{}
   \label{subfig:noise_shape}
  \end{subfigure}
  \hfill
  \begin{subfigure}[b]{0.42\textwidth}
   \includegraphics[width=\textwidth]{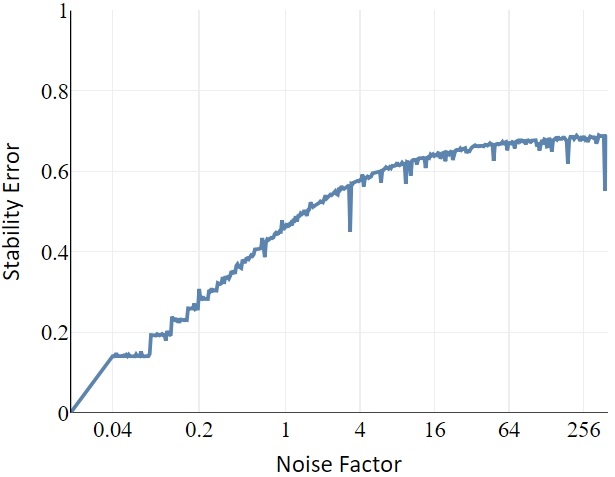}
   \caption{}
   \label{subfig:noise_stab}
  \end{subfigure}
  \caption{The relationship between noise and stability error. (a) When a noise of the same magnitude is introduced to embeddings of varying shapes ($n \times d$), stability error values are quite stable; demonstrating the measure's robustness against number of objects and embedding size. (b) Stability error reflects the magnitude of the introduced noise as more noise is assumed to reflect more dramatic structural changes in the dynamic system.}
  \label{fig:noise_stability}
\end{figure}

As is the case with other measures, we expect stability error to be robust against different embedding shapes. To test this, we create initial matrices following the procedure we employed earlier, change their structure via the introduction of noise with the noise factor fixed at $1$, and observe the stability error between the pairs of embeddings. Figure \ref{subfig:noise_shape} visualizes the result of this experiment. Except for the regime characterized by very low $n/d$ values, the stability error is steady across the number of objects and embedding length. This low-$n/d$ regime is not very relevant in real-world cases since representation learning aims to learn embeddings satisfying $d<<n$. Moreover, as soon as the number of nodes modestly increases to above $100$, it already reaches the robust region even if $d$ is relatively large at $32$. In light of these results, we see that our stability measure is robust to the shape of embeddings under realistic scenarios. Following this, using a large number of initial matrices of different shapes as usual, we plot stability error against the noise factor in Figure \ref{subfig:noise_stab}. The lack of large fluctuations also confirms the robustness of stability error to embedding shapes. On the other hand, we observe that stability error does not reach its maximum value but diminishes around $0.7$. This is because the upper bound of stability error is produced only under very extreme cases that are practically improbable to replicate with random embeddings and noise. In addition to the robustness result, Figure \ref{subfig:noise_stab} demonstrates the direct relationship between the extent of change in the embedding structure and values of the stability error.

\begin{figure}[!h]
  \centering
  \begin{subfigure}[b]{0.333\textwidth}
   \includegraphics[width=\textwidth]{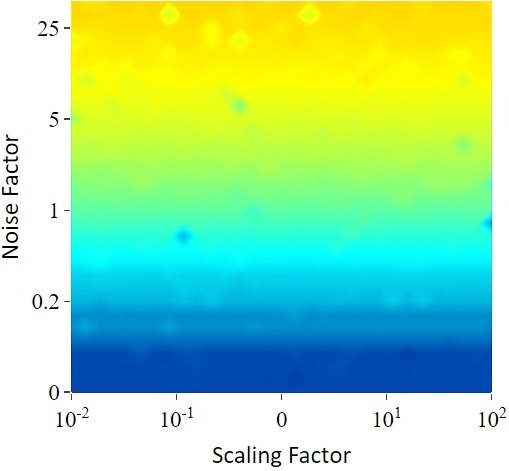}
   \caption{}
   \label{subfig:scale}
  \end{subfigure}
  \hfill
  \begin{subfigure}[b]{0.299\textwidth}
   \includegraphics[width=\textwidth]{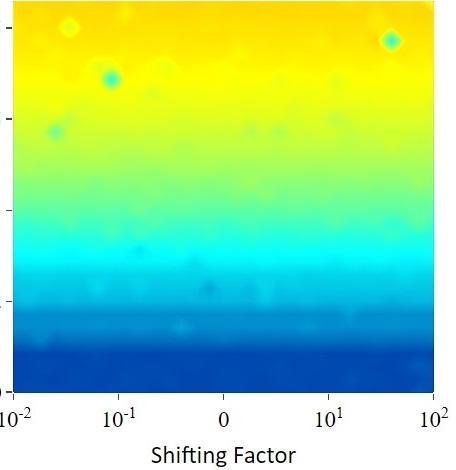}
   \caption{}
   \label{subfig:shift}
  \end{subfigure}
  \hfill
  \begin{subfigure}[b]{0.353\textwidth}
   \includegraphics[width=\textwidth]{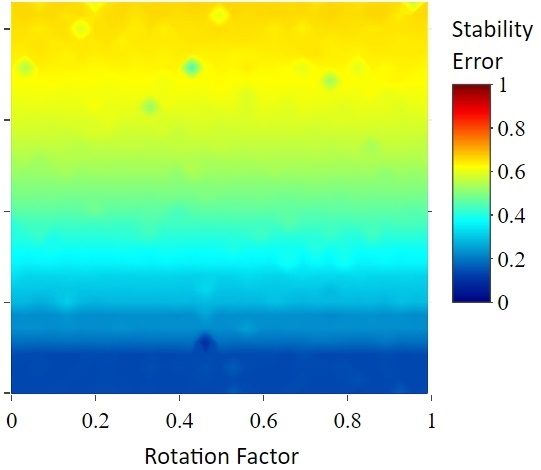}
   \caption{}
   \label{subfig:rotation}
  \end{subfigure}
  \caption{Combined effect of noise and scaling, shifting, or rotation on stability error. The subfigures share the y-axis and the colorbar. (a) Scaling and noise factors, and corresponding stability errors. (b) Shifting and noise factors, and corresponding stability errors. (c) Rotation and noise factors, and corresponding stability errors. Overall, stability error reflects the magnitude of noise; and scaling, shifting, and rotation have no effect on it.}
  \label{fig:transform_noise_stability}
\end{figure}

In the presence of both structural changes achieved by random noise walks and other transformations that do not alter the relative structure, we expect stability error to be independent of such linear transformations. To test and demonstrate this, a large number of initial matrices are created with the usual procedure, and the linear transformations and noise are introduced with varying magnitudes. Figure \ref{fig:transform_noise_stability} presents the results of these experiments for each noise and linear transformation pair. As shown by the figure, stability error is linked to the extent of changes introduced to the relative structure of objects and completely independent from transformations that do not impact the relative structure.

Finally, we investigate behaviors of the scale, translation, and rotation errors against their respective linear transformations and the noise added by the random walk procedure. Figure \ref{subfig:scale} shows that the scale error, in addition to the magnitude of the applied scaling, is affected also by the magnitude of the noise. Specifically, as we contract the embedding (i.e., scaling factor $< 1$), the scale error does not increase as fast if a large magnitude of the noise is introduced as well. The reason for this is rather simple: as objects take more random steps in lengthier walks, they spread further than their original boundaries and the radius of the embeddings slowly increases. Such an increase in the radius plays a balancing role against contraction, resulting in the observed behavior in the figure. The tendency of random walks to increase the radius also causes the behavior observed in Figure \ref{subfig:shift}. The increasing radius increases the normalization term that is used in the translation error computation (i.e. $r^t + r^{t+1}$ in Equation \ref{tnorm}). As a direct consequence, under very large random walks, translation error gets reduced. Overall, such behavior resulting from the increasing radius is due to the way we model noise in our synthetic experiments rather than the characteristics or properties of our measure and is visible conspicuously only in very lengthy random noise walks.

Very lengthy random noise walks, as shown in Figure \ref{subfig:rotation}, transforms any embedding matrix into a random-like matrix. In the regime characterized by large noise factors, the rotation error converges to a large value even if the initial rotation factors are different. This value is not the upper bound of the rotation error since its worst value is obtained with $180^{\circ}$ rotations and an additional reflection for odd dimensions (i.e., with the negative identity matrix) instead of a random matrix. In the regime characterized by moderate and more realistic noise levels, rotation error is able to distinguish between the noise and original structure and produces values in line with the respective values of the rotation factor. Keeping in mind that a noise factor of $1$ produces a random noise with the same magnitude as the original embedding, the result also demonstrates the capability of our rotation error measure to capture the rotation differences even under considerable amounts of noise.

Here, we should note again that alignment is relevant for the cases where it is possible to find a matching structural pattern between two embeddings. A large stability error indicates that two embeddings are structurally so dissimilar that it is not meaningful to investigate alignment and interpret the results of other error measures. Therefore, the behavior of scale, translation, and rotation errors are not relevant, significant, or practically useful when the length of random noise walk is too large that the introduced noise suppresses the original structure.

\begin{figure}[!h]
  \centering
  \begin{subfigure}[b]{0.325\textwidth}
   \includegraphics[width=\textwidth]{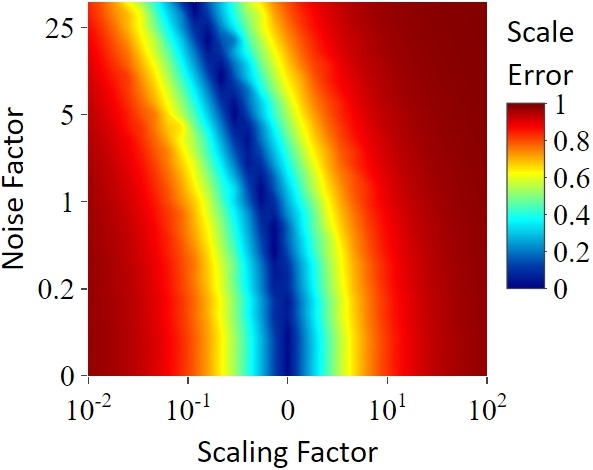}
   \caption{}
   \label{subfig:scale}
  \end{subfigure}
  \hfill
  \begin{subfigure}[b]{0.325\textwidth}
   \includegraphics[width=\textwidth]{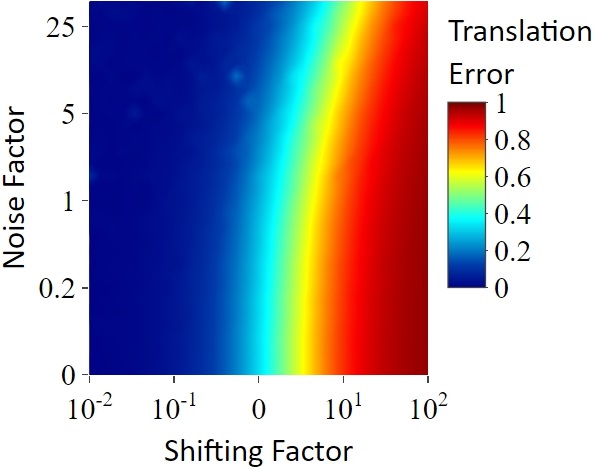}
   \caption{}
   \label{subfig:shift}
  \end{subfigure}
  \hfill
  \begin{subfigure}[b]{0.325\textwidth}
   \includegraphics[width=\textwidth]{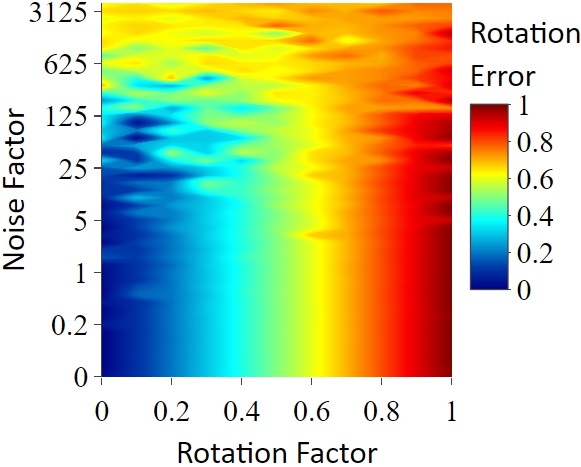}
   \caption{}
   \label{subfig:rotation}
  \end{subfigure}
  \caption{Combined effect of noise and scaling, shifting, or rotation on scale, translation, and rotation errors. (a) Scale error reflects the magnitude of scaling. It is also affected by the magnitude of noise of due to the increasing radius. (b) Translation error reflects the magnitude of shifting. It is also affected by the magnitude of noise of due to the increasing radius. (c) Rotation error reflects the magnitude of rotation. It is also affected by the magnitude of noise since very long random noise walks loses the information in the original embedding.}
  \label{fig:transform_noise_transform}
\end{figure} 

In summary, the proposed error measures for stability, scale, translation, and rotation are able to capture the existence and extent of differences introduced by respective structural changes or transformations. They are robust to the different number of objects and embedding vector lengths, and each of them directly and only reflects the changes that it is supposed to measure. These four measures together are fit for use in real-world cases to investigate the existence and degree of stability and alignment and realign embeddings that are stable but misaligned.

\section{Experiments on real-world networks}

In this section, we conduct real-world experiments on embeddings generated for seven real-world datasets using 13 network representation learning methods. First, we measure alignment and stability errors in these embeddings. Then, using the original and aligned versions of these embeddings as input features, we build node classification models and observe their performance on the unseen embedding of the next timestep. In another experiment, we introduce different degrees of misalignment and observe the change in node classification task performance.

\textbf{Datasets.} Using network dataset repositories \cite{netzschleuder, sociopatterns} and other resources, we identified seven real-world dynamic networks with fixed or evolving metadata for nodes. For each network, we take appropriate preprocessing steps to ensure that the node set is fixed and all networks are connected over all timesteps. Information on the datasets is provided below and a summary of the final network datasets is available in Table \ref{tab:datasets}.

\begin{itemize}[leftmargin=*]

 \item \textit{dutch-school} network \cite{dutchschool} denotes the friendship network among freshmen students at a secondary school in Netherlands. The data were collected 4 times with 3-month intervals in 2003 and 2004. Each directed link denotes whether a student considers the other one as a friend or not. Fixed node labels are sex, age, ethnicity, and religion. The evolving node label is delinquency count which we transform into a binary variable showing whether a student has more than one delinquency for the last interval.
 
 \item \textit{freshmen} network \cite{freshmen} denotes the friendship network among freshmen students at the Sociology department of a university in the Netherlands. The data were collected five times in 1998 and 1999. Each directed link weight denotes the strength of the friendship from the perspective of one student. Fixed node labels gender, program, and smoking level. Smoking level is transformed into whether a student is a regular smoker.
 
 \item \textit{sp-hospital} network \cite{sphospital} denotes the face-to-face contact data between patients and healthcare workers at a hospital in France from a Monday noon to the next Friday noon in 2010. We process the network such that each link weight represents the total duration of interaction between a pair of people in 24 hours; obtaining 4 timesteps. The fixed node label is the job title or being a patient.
 
 \item \textit{sp-primary-school} network \cite{spprimaryschool} denotes the face-to-face contact data on 2 consecutive days at a primary school in France in 2009. Link weights represent the duration of respective face-to-face interaction over a single day. Fixed node labels are class and gender.
 
 \item \textit{sp-high-school} network \cite{sphighschool} denotes the face-to-face contact data between students in a high school in France over 5 days in 2013. We process the network such that each link weight represents the total duration of interaction between a pair of students in one day; obtaining 5 timesteps. Fixed node labels are class and gender.
  
 \item \textit{aminer} network \cite{aminer} denotes the coauthorship relations between researcher. We processed the network such that each edge weight represents the number of coauthored work for a pair of researchers over a 3-year period. There is a total of four 3-year periods, spanning 12 years in total. The evolving node label is the research field.
 
 \item \textit{yahoo} network \cite{yahoo} denotes the communication between a sample of users of the Yahoo! Messenger over 4 weeks. The dataset is provided to us through The Yahoo Webscope Program. We process the data such that each directed link weight represents the number of days a user sent a message to the other in a given week; obtaining 4 timesteps. The evolving node label is a more general and more specific location area of the user, obtained from the first one and two letters of the most frequent postcodes in a given week.

\end{itemize}

\begin{table}[!h]
\footnotesize
\centering
\setlength{\tabcolsep}{0.1em} 
\renewcommand{\arraystretch}{0.75}
\caption{Datasets}
\label{tab:datasets}
\begin{tabular}{|l|l|l|l|l|l|l|}
\hline
\textbf{Name} & \textbf{t} & \textbf{n} & \textbf{Directed} & \textbf{Weighted} & \textbf{Fixed metadata} & \textbf{Evolving metadata} \\ \hline
dutch-school  & 4 & 25 & TRUE & FALSE & sex, age, ethnicity, religion  & delinquency count \\ \hline
freshmen & 4 & 30 & TRUE & TRUE & gender, age, program, smoking & n/a    \\ \hline
sp-hospital  & 2 & 51 & FALSE & TRUE & job         & n/a    \\ \hline
sp-primary-school & 2 & 232 & FALSE & TRUE & class, gender       & n/a    \\ \hline
sp-high-school  & 5 & 244 & FALSE & TRUE & gender, class       & n/a    \\ \hline
aminer    & 4 & 10565 & FALSE & TRUE & n/a         & research field \\ \hline
yahoo & 4 & 46049 & TRUE & TRUE & n/a         & postcode   \\ \hline
\end{tabular}
\end{table}

We utilize a variety of static and dynamic network representation learning methods. As static methods, we employ random-walk-based methods such as \textit{node2vec}\cite{grover2016node2vec}, \textit{DeepWalk}\cite{perozzi2014deepwalk}; matrix-factorization-based methods such as \textit{GF}, \textit{LAP}\cite{belkin2002laplacian}, \textit{HOPE}\cite{ou2016asymmetric}; dimension reduction-based method such as \textit{LLE}\cite{RoweisLLE}; neural-network-based method such as \textit{SDNE}\cite{wang2016structural}; and large graph embedding methods such as \textit{LINE}\cite{tang2015line}. These methods which are originally designed for static networks are applied to generate embeddings for static snapshots of networks at each timestep without any information sharing between different timesteps. As dynamic methods, we employ \textit{DynamicTriad}\cite{zhou2018dynamic}, \textit{GloDyNE}\cite{hou2020glodyne}, and \textit{temporalnode2vec}\cite{haddad2019temporalnode2vec}. In addition, we employ the temporalized versions of \textit{DeepWalk} and \textit{node2vec} using the following process. We create $t$ versions of each node for each timestep and concatenate the edge list. In this way, we obtain $t$ disconnected components in a single graph where each component corresponds to a timestep. Then, to ensure information sharing between different timesteps, we create links between the different versions of the same nodes in consecutive timesteps. On this network, we can apply any static representation learning method and obtain a separate embedding vector for each node at each timestep. Due to space limitations, we utilize only \textit{node2vec} and \textit{DeepWalk} to obtain their temporalized versions \textit{tmpN2V} and \textit{tmpDW} respectively. Lastly, for embeddings methods that require hyperparameters, we choose values based on popular default values available in the relevant literature\footnote{Since our objective does not lie in methods comparison, we do not test different hyperparameters}.

\begin{figure}[!h]
\centering
\includegraphics[width=\textwidth]{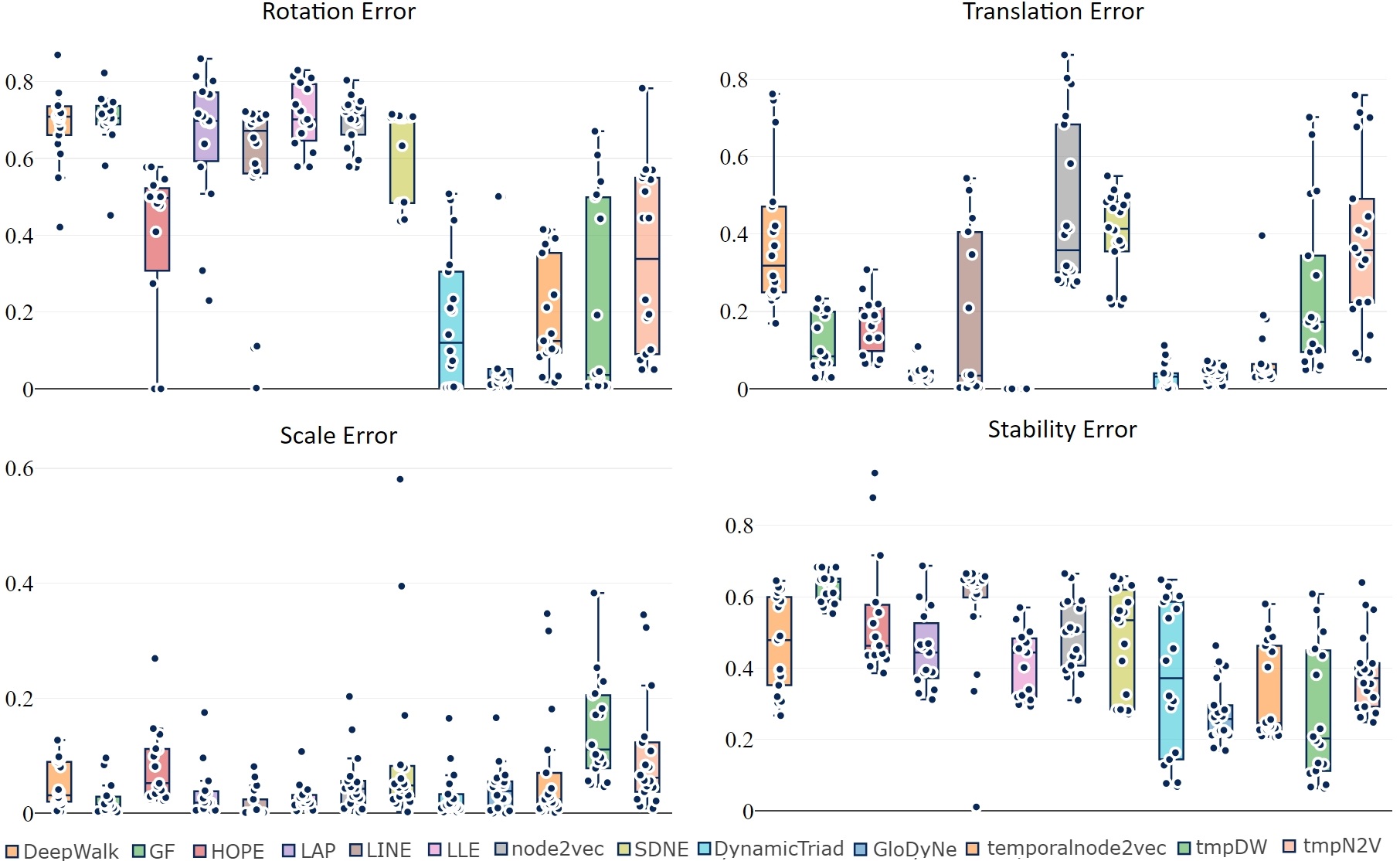}
\caption{Alignment and stability errors produced by embedding methods. Methods are ordered based on whether they are static or dynamic. Boxplot whiskers are up to 1.5 interquartile range away from Q1 and Q3. Individual data points are shown with circle markers.}
\label{fig:rw_errors}
\end{figure}

Figure \ref{fig:rw_errors} demonstrates rotation, translation, scale, and stability errors produced by each method over all networks and consecutive timesteps. With respect to rotation, we observe that the static methods produce larger errors than the dynamic methods. However, the rotation errors produced by dynamic methods are not insignificant either, with a possible exception of \textit{GloDyNE} which produce only very small rotation errors. With respect to the translation error, random-walk-based methods produce larger errors as well as \textit{SDNE} and \textit{LINE}. Except for the temporalized versions, dynamic methods usually produce lower translation errors than the static ones with a very notable exception of \textit{LLE}. With respect to scale error, we observe that most methods usually produce very small errors whereas temporalized methods produce relatively larger errors. With respect to stability error, as expected, we do not observe very meaningful differences between static and dynamic methods. Still, dynamic methods usually produce smaller stability errors possibly due to the information sharing between timesteps which explicitly or implicitly drives embeddings to be closer between consecutive timesteps.

Overall, our empirical analysis using popular default hyperparameter values for all methods shows that static methods almost always produce large rotation errors and often produce large translation errors. This is expected since the embeddings are learned independently in different timesteps. However, the alignment errors (i.e., rotation and translation errors) produced by dynamic methods are often not trivial either. This result shows that even dynamic methods in the literature do not always guarantee embedding alignment between timesteps.

Next, we perform a network inference task using the available fixed or evolving label information. For each network embedding and node label, we train a Support Vector Machine (SVM) using the embedding matrix at time $t$ as input features and labels as the output variable. Then, using the model, we predict the labels using the embedding matrix at time $t+1$. Afterward, we find optimal transformations to ensure alignment and align all embedding matrices over timesteps. After ensuring alignment (i.e., with rotation and translation errors now equal to zero), we perform the same experiment. As a result of this experiment, we obtain classification accuracy scores separately for the original and aligned versions of the embeddings on the unseen test datasets of embeddings at $t+1$.

Not all labels can be predicted by the network data. For instance, in a setting with perfect gender mixing, there is no information available in the network structure to predict gender. In order to filter out cases with no meaningful learning for the classification tasks, we compare the accuracy scores obtained using aligned embeddings to a dummy classifier. The dummy classifier always outputs the most frequent label as the prediction for all nodes. With $a$ and $b$ respectively denoting the accuracy scores of the learned model and the dummy classifier, we filter out the cases if the following condition is not satisfied: $a > 0.25(1-b) + b$. In other words, the model should capture the \%25 of the information that could not be captured by the dummy classifier. 

\begin{figure}[!h]
 \centering
 \begin{subfigure}[b]{0.385\textwidth}
  \includegraphics[width=\textwidth]{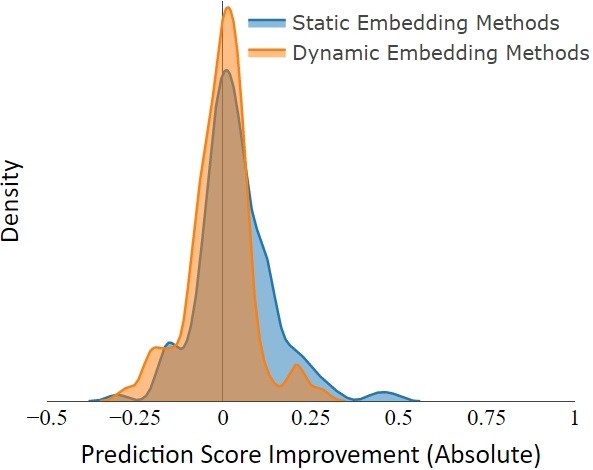}
  \caption{}
 \label{fig:dist_improv_a}
 \end{subfigure}
 \hspace{5mm}
 \begin{subfigure}[b]{0.385\textwidth}
  \includegraphics[width=\textwidth]{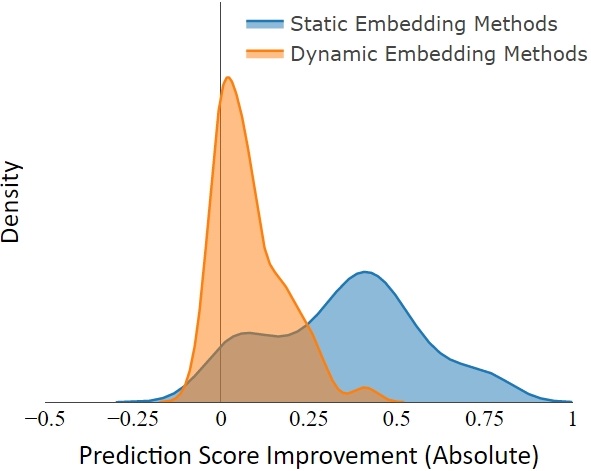}
  \caption{}
 \label{fig:dist_improv_b}
 \end{subfigure}
 \caption{Distribution of prediction score improvement after alignment. Subfigures share the same y-axis. (a) Cases that are filtered out based on the comparison with the dummy classifier. The improvement values are roughly distributed normally around 0, both for the static and dynamic methods. (b) Cases that are retained based on the comparison with the dummy classifier. The improvement values are mostly positive; and are greater on average for the static methods in comparison to the dynamic methods.}
 \label{fig:dist_improv}
\end{figure}

Figure \ref{fig:dist_improv} demonstrates the absolute difference in classification accuracy scores before and after aligning the embeddings, separately for dynamic and static network embedding methods. Figure \ref{fig:dist_improv_a} shows the cases that are filtered out. Confirming our intuitions, there are no significant improvements in classification accuracy in filtered out cases and the values are approximately normally distributed around 0. This is in line with our expectation that if the model is not better at the classification task than a dummy classifier, it is not appropriate to expect improvements after alignment. In the rest of the analysis in this section, we only consider models that are sufficiently better than the dummy classifier. Figure \ref{fig:dist_improv_b} contains the cases satisfying the above-stated condition. We observe that, in most cases, there is an increase in the classification accuracy after alignment. The small number of negative values are also small in magnitude and are exceptions that may be attributed to some random effects. When alignment is applied as post-processing to the embeddings from static methods, we observe consistent improvement in classification accuracy reaching to \%90. In the case of dynamic networks, we still observe very meaningful and consistent performance increases in classification accuracy, with the difference reaching to more than \%40. Hence, we can conclude that not only the static but also the dynamic embedding methods benefit from alignment.

\begin{table}[!h]
\footnotesize
\centering
\setlength{\tabcolsep}{0.1em} 
\renewcommand{\arraystretch}{0.75}
\caption{Prediction accuracy before and after alignment}
\label{tab:rw_improv_align}

\begin{tabular}{llllllllllllllllllllllllllll}
\cline{2-28}
\multicolumn{1}{l|}{} &
 \multicolumn{8}{c|}{sp-high-school} &
 \multicolumn{6}{c|}{aminer} &
 \multicolumn{12}{c|}{yahoo} &
 \multicolumn{1}{l|}{\multirow{4}{*}{\textit{O-A}}} \\ \cline{2-27}
\multicolumn{1}{l|}{} &
 \multicolumn{8}{c|}{class} &
 \multicolumn{6}{c|}{field} &
 \multicolumn{6}{c|}{general area} &
 \multicolumn{6}{c|}{specific area} &
 \multicolumn{1}{l|}{} \\ \cline{2-27}
\multicolumn{1}{l|}{} &
 \multicolumn{2}{c|}{0-1} &
 \multicolumn{2}{c|}{1-2} &
 \multicolumn{2}{c|}{2-3} &
 \multicolumn{2}{c|}{3-4} &
 \multicolumn{2}{c|}{0-1} &
 \multicolumn{2}{c|}{1-2} &
 \multicolumn{2}{c|}{2-3} &
 \multicolumn{2}{c|}{0-1} &
 \multicolumn{2}{c|}{1-2} &
 \multicolumn{2}{c|}{2-3} &
 \multicolumn{2}{c|}{0-1} &
 \multicolumn{2}{c|}{1-2} &
 \multicolumn{2}{c|}{2-3} &
 \multicolumn{1}{l|}{} \\ \cline{2-27}
\multicolumn{1}{l|}{} &
 \multicolumn{1}{c|}{O} &
 \multicolumn{1}{c|}{A} &
 \multicolumn{1}{c|}{O} &
 \multicolumn{1}{c|}{A} &
 \multicolumn{1}{c|}{O} &
 \multicolumn{1}{c|}{A} &
 \multicolumn{1}{c|}{O} &
 \multicolumn{1}{c|}{A} &
 \multicolumn{1}{c|}{O} &
 \multicolumn{1}{c|}{A} &
 \multicolumn{1}{c|}{O} &
 \multicolumn{1}{c|}{A} &
 \multicolumn{1}{c|}{O} &
 \multicolumn{1}{c|}{A} &
 \multicolumn{1}{c|}{O} &
 \multicolumn{1}{c|}{A} &
 \multicolumn{1}{c|}{O} &
 \multicolumn{1}{c|}{A} &
 \multicolumn{1}{c|}{O} &
 \multicolumn{1}{c|}{A} &
 \multicolumn{1}{c|}{O} &
 \multicolumn{1}{c|}{A} &
 \multicolumn{1}{c|}{O} &
 \multicolumn{1}{c|}{A} &
 \multicolumn{1}{c|}{O} &
 \multicolumn{1}{l|}{A} &
 \multicolumn{1}{l|}{} \\ \cline{2-28} 
 &
 &
 &
 &
 &
 &
 &
 &
 &
 &
 &
 &
 &
 &
 &
 &
 &
 &
 &
 &
 &
 &
 &
 &
 &
 &
 &
  \vspace{-1.5mm}
 \\ \hline
\multicolumn{1}{|l|}{DeepWalk} &
 \multicolumn{1}{l|}{.01} &
 \multicolumn{1}{l|}{.55} &
 \multicolumn{1}{l|}{.13} &
 \multicolumn{1}{l|}{.66} &
 \multicolumn{1}{l|}{.12} &
 \multicolumn{1}{l|}{.74} &
 \multicolumn{1}{l|}{.09} &
 \multicolumn{1}{l|}{.90} &
 \multicolumn{1}{l|}{.34} &
 \multicolumn{1}{l|}{.68} &
 \multicolumn{1}{l|}{.24} &
 \multicolumn{1}{l|}{.69} &
 \multicolumn{1}{l|}{.35} &
 \multicolumn{1}{l|}{.69} &
 \multicolumn{1}{l|}{.17} &
 \multicolumn{1}{l|}{.63} &
 \multicolumn{1}{l|}{.17} &
 \multicolumn{1}{l|}{.64} &
 \multicolumn{1}{l|}{.18} &
 \multicolumn{1}{l|}{.64} &
 \multicolumn{1}{l|}{.03} &
 \multicolumn{1}{l|}{.43} &
 \multicolumn{1}{l|}{.03} &
 \multicolumn{1}{l|}{.44} &
 \multicolumn{1}{l|}{.04} &
 \multicolumn{1}{l|}{.44} &
 \multicolumn{1}{l|}{\textit{.48}} \\ \hline
\multicolumn{1}{|l|}{GF} &
 \multicolumn{1}{l|}{.12} &
 \multicolumn{1}{l|}{.16} &
 \multicolumn{1}{l|}{.09} &
 \multicolumn{1}{l|}{.11} &
 \multicolumn{1}{l|}{.12} &
 \multicolumn{1}{l|}{.13} &
 \multicolumn{1}{l|}{.11} &
 \multicolumn{1}{l|}{.13} &
 \multicolumn{1}{l|}{.26} &
 \multicolumn{1}{l|}{.58} &
 \multicolumn{1}{l|}{.27} &
 \multicolumn{1}{l|}{.56} &
 \multicolumn{1}{l|}{.26} &
 \multicolumn{1}{l|}{.60} &
 &
 &
 &
 &
 &
 &
 &
 &
 &
 &
 &
 \multicolumn{1}{l|}{} &
 \multicolumn{1}{l|}{\textit{.15}} \\ \cline{1-15} \cline{28-28} 
\multicolumn{1}{|l|}{HOPE} &
 \multicolumn{1}{l|}{.12} &
 \multicolumn{1}{l|}{.11} &
 \multicolumn{1}{l|}{.14} &
 \multicolumn{1}{l|}{.13} &
 \multicolumn{1}{l|}{.13} &
 \multicolumn{1}{l|}{.15} &
 \multicolumn{1}{l|}{.09} &
 \multicolumn{1}{l|}{.15} &
 \multicolumn{1}{l|}{.36} &
 \multicolumn{1}{l|}{.36} &
 \multicolumn{1}{l|}{.26} &
 \multicolumn{1}{l|}{.27} &
 \multicolumn{1}{l|}{.13} &
 \multicolumn{1}{l|}{.21} &
 &
 &
 &
 &
 &
 &
 &
 &
 &
 &
 &
 \multicolumn{1}{l|}{} &
 \multicolumn{1}{l|}{\textit{.02}} \\ \cline{1-15} \cline{28-28} 
\multicolumn{1}{|l|}{LAP} &
 \multicolumn{1}{l|}{.14} &
 \multicolumn{1}{l|}{.80} &
 \multicolumn{1}{l|}{.24} &
 \multicolumn{1}{l|}{.65} &
 \multicolumn{1}{l|}{.01} &
 \multicolumn{1}{l|}{.73} &
 \multicolumn{1}{l|}{.07} &
 \multicolumn{1}{l|}{.80} &
 \multicolumn{1}{l|}{.21} &
 \multicolumn{1}{l|}{.46} &
 \multicolumn{1}{l|}{.26} &
 \multicolumn{1}{l|}{.37} &
 \multicolumn{1}{l|}{.06} &
 \multicolumn{1}{l|}{.26} &
 &
 &
 &
 &
 &
 &
 &
 &
 &
 &
 &
 \multicolumn{1}{l|}{} &
 \multicolumn{1}{l|}{\textit{.44}} \\ \hline
\multicolumn{1}{|l|}{LINE} &
 \multicolumn{1}{l|}{.11} &
 \multicolumn{1}{l|}{.14} &
 \multicolumn{1}{l|}{.22} &
 \multicolumn{1}{l|}{.17} &
 \multicolumn{1}{l|}{.14} &
 \multicolumn{1}{l|}{.12} &
 \multicolumn{1}{l|}{.16} &
 \multicolumn{1}{l|}{.13} &
 \multicolumn{1}{l|}{.33} &
 \multicolumn{1}{l|}{.51} &
 \multicolumn{1}{l|}{.27} &
 \multicolumn{1}{l|}{.48} &
 \multicolumn{1}{l|}{.15} &
 \multicolumn{1}{l|}{.52} &
 \multicolumn{1}{l|}{.16} &
 \multicolumn{1}{l|}{.59} &
 \multicolumn{1}{l|}{.17} &
 \multicolumn{1}{l|}{.62} &
 \multicolumn{1}{l|}{.21} &
 \multicolumn{1}{l|}{.59} &
 \multicolumn{1}{l|}{.03} &
 \multicolumn{1}{l|}{.51} &
 \multicolumn{1}{l|}{.03} &
 \multicolumn{1}{l|}{.54} &
 \multicolumn{1}{l|}{.05} &
 \multicolumn{1}{l|}{.50} &
 \multicolumn{1}{l|}{\textit{.26}} \\ \hline
\multicolumn{1}{|l|}{LLE} &
 \multicolumn{1}{l|}{.23} &
 \multicolumn{1}{l|}{.78} &
 \multicolumn{1}{l|}{.01} &
 \multicolumn{1}{l|}{.65} &
 \multicolumn{1}{l|}{.24} &
 \multicolumn{1}{l|}{.74} &
 \multicolumn{1}{l|}{.02} &
 \multicolumn{1}{l|}{.81} &
 \multicolumn{1}{l|}{.31} &
 \multicolumn{1}{l|}{.51} &
 \multicolumn{1}{l|}{.29} &
 \multicolumn{1}{l|}{.43} &
 \multicolumn{1}{l|}{.25} &
 \multicolumn{1}{l|}{.39} &
 &
 &
 &
 &
 &
 &
 &
 &
 &
 &
 &
 \multicolumn{1}{l|}{} &
 \multicolumn{1}{l|}{\textit{.42}} \\ \hline
\multicolumn{1}{|l|}{node2vec} &
 \multicolumn{1}{l|}{.08} &
 \multicolumn{1}{l|}{.86} &
 \multicolumn{1}{l|}{.07} &
 \multicolumn{1}{l|}{.52} &
 \multicolumn{1}{l|}{.02} &
 \multicolumn{1}{l|}{.74} &
 \multicolumn{1}{l|}{.01} &
 \multicolumn{1}{l|}{.60} &
 \multicolumn{1}{l|}{.24} &
 \multicolumn{1}{l|}{.70} &
 \multicolumn{1}{l|}{.25} &
 \multicolumn{1}{l|}{.70} &
 \multicolumn{1}{l|}{.22} &
 \multicolumn{1}{l|}{.71} &
 \multicolumn{1}{l|}{.16} &
 \multicolumn{1}{l|}{.69} &
 \multicolumn{1}{l|}{.21} &
 \multicolumn{1}{l|}{.70} &
 \multicolumn{1}{l|}{.18} &
 \multicolumn{1}{l|}{.70} &
 \multicolumn{1}{l|}{.02} &
 \multicolumn{1}{l|}{.56} &
 \multicolumn{1}{l|}{.02} &
 \multicolumn{1}{l|}{.59} &
 \multicolumn{1}{l|}{.03} &
 \multicolumn{1}{l|}{.58} &
 \multicolumn{1}{l|}{\textit{.55}} \\ \hline
\multicolumn{1}{|l|}{SDNE} &
 \multicolumn{1}{l|}{.12} &
 \multicolumn{1}{l|}{.46} &
 \multicolumn{1}{l|}{.18} &
 \multicolumn{1}{l|}{.55} &
 \multicolumn{1}{l|}{.08} &
 \multicolumn{1}{l|}{.50} &
 \multicolumn{1}{l|}{.10} &
 \multicolumn{1}{l|}{.51} &
 \multicolumn{1}{l|}{.33} &
 \multicolumn{1}{l|}{.38} &
 \multicolumn{1}{l|}{.19} &
 \multicolumn{1}{l|}{.31} &
 \multicolumn{1}{l|}{.16} &
 \multicolumn{1}{l|}{.28} &
 \multicolumn{1}{l|}{.20} &
 \multicolumn{1}{l|}{.31} &
 \multicolumn{1}{l|}{.21} &
 \multicolumn{1}{l|}{.35} &
 \multicolumn{1}{l|}{.19} &
 \multicolumn{1}{l|}{.39} &
 \multicolumn{1}{l|}{.05} &
 \multicolumn{1}{l|}{.16} &
 \multicolumn{1}{l|}{.04} &
 \multicolumn{1}{l|}{.21} &
 \multicolumn{1}{l|}{.04} &
 \multicolumn{1}{l|}{.26} &
 \multicolumn{1}{l|}{\textit{.21}} \\ \hline
\multicolumn{1}{|l|}{\textit{O-A}} &
 \multicolumn{2}{c|}{\textit{.37}} &
 \multicolumn{2}{c|}{\textit{.30}} &
 \multicolumn{2}{c|}{\textit{.37}} &
 \multicolumn{2}{c|}{\textit{.42}} &
 \multicolumn{2}{c|}{\textit{.23}} &
 \multicolumn{2}{c|}{\textit{.22}} &
 \multicolumn{2}{c|}{\textit{.26}} &
 \multicolumn{2}{c|}{\textit{.38}} &
 \multicolumn{2}{c|}{\textit{.39}} &
 \multicolumn{2}{c|}{\textit{.39}} &
 \multicolumn{2}{c|}{\textit{.38}} &
 \multicolumn{2}{c|}{\textit{.41}} &
 \multicolumn{2}{c|}{\textit{.41}} &
 \textit{} \\ \cline{1-27}
 &
 &
 &
 &
 &
 &
 &
 &
 &
 &
 &
 &
 &
 &
 &
 &
 &
 &
 &
 &
 &
 &
 &
 &
 &
 &
 &
 \vspace{-1.5mm}
 \textit{} \\ \hline
\multicolumn{1}{|l|}{Dyn'Triad} &
 \multicolumn{1}{l|}{.18} &
 \multicolumn{1}{l|}{.24} &
 \multicolumn{1}{l|}{.14} &
 \multicolumn{1}{l|}{.14} &
 \multicolumn{1}{l|}{.17} &
 \multicolumn{1}{l|}{.21} &
 \multicolumn{1}{l|}{.19} &
 \multicolumn{1}{l|}{.19} &
 \multicolumn{1}{l|}{.52} &
 \multicolumn{1}{l|}{.55} &
 \multicolumn{1}{l|}{.49} &
 \multicolumn{1}{l|}{.53} &
 \multicolumn{1}{l|}{.53} &
 \multicolumn{1}{l|}{.57} &
 \multicolumn{1}{l|}{.68} &
 \multicolumn{1}{l|}{.68} &
 \multicolumn{1}{l|}{.75} &
 \multicolumn{1}{l|}{.75} &
 \multicolumn{1}{l|}{.72} &
 \multicolumn{1}{l|}{.72} &
 \multicolumn{1}{l|}{.61} &
 \multicolumn{1}{l|}{.61} &
 \multicolumn{1}{l|}{.69} &
 \multicolumn{1}{l|}{.69} &
 \multicolumn{1}{l|}{.65} &
 \multicolumn{1}{l|}{.65} &
 \multicolumn{1}{l|}{\textit{.02}} \\ \hline
\multicolumn{1}{|l|}{GloDyNE} &
 \multicolumn{1}{l|}{.96} &
 \multicolumn{1}{l|}{.96} &
 \multicolumn{1}{l|}{.92} &
 \multicolumn{1}{l|}{.91} &
 \multicolumn{1}{l|}{.92} &
 \multicolumn{1}{l|}{.91} &
 \multicolumn{1}{l|}{.93} &
 \multicolumn{1}{l|}{.95} &
 \multicolumn{1}{l|}{.65} &
 \multicolumn{1}{l|}{.70} &
 \multicolumn{1}{l|}{.66} &
 \multicolumn{1}{l|}{.69} &
 \multicolumn{1}{l|}{.62} &
 \multicolumn{1}{l|}{.70} &
 \multicolumn{1}{l|}{.62} &
 \multicolumn{1}{l|}{.71} &
 \multicolumn{1}{l|}{.63} &
 \multicolumn{1}{l|}{.74} &
 \multicolumn{1}{l|}{.63} &
 \multicolumn{1}{l|}{.73} &
 \multicolumn{1}{l|}{.42} &
 \multicolumn{1}{l|}{.60} &
 \multicolumn{1}{l|}{.45} &
 \multicolumn{1}{l|}{.65} &
 \multicolumn{1}{l|}{.45} &
 \multicolumn{1}{l|}{.63} &
 \multicolumn{1}{l|}{\textit{.08}} \\ \hline
\multicolumn{1}{|l|}{t'node2vec} &
 \multicolumn{1}{l|}{.89} &
 \multicolumn{1}{l|}{.93} &
 \multicolumn{1}{l|}{.56} &
 \multicolumn{1}{l|}{.75} &
 \multicolumn{1}{l|}{.77} &
 \multicolumn{1}{l|}{.85} &
 \multicolumn{1}{l|}{.90} &
 \multicolumn{1}{l|}{.87} &
 \multicolumn{1}{l|}{.49} &
 \multicolumn{1}{l|}{.58} &
 \multicolumn{1}{l|}{.39} &
 \multicolumn{1}{l|}{.54} &
 \multicolumn{1}{l|}{.34} &
 \multicolumn{1}{l|}{.44} &
 \multicolumn{1}{l|}{.33} &
 \multicolumn{1}{l|}{.36} &
 \multicolumn{1}{l|}{.33} &
 \multicolumn{1}{l|}{.37} &
 \multicolumn{1}{l|}{.33} &
 \multicolumn{1}{l|}{.37} &
 \multicolumn{1}{l|}{.16} &
 \multicolumn{1}{l|}{.20} &
 \multicolumn{1}{l|}{.17} &
 \multicolumn{1}{l|}{.21} &
 \multicolumn{1}{l|}{.17} &
 \multicolumn{1}{l|}{.20} &
 \multicolumn{1}{l|}{\textit{.06}} \\ \hline
\multicolumn{1}{|l|}{tmpDW} &
 \multicolumn{1}{l|}{.59} &
 \multicolumn{1}{l|}{.83} &
 \multicolumn{1}{l|}{.48} &
 \multicolumn{1}{l|}{.89} &
 \multicolumn{1}{l|}{.52} &
 \multicolumn{1}{l|}{.72} &
 \multicolumn{1}{l|}{.52} &
 \multicolumn{1}{l|}{.31} &
 \multicolumn{1}{l|}{.66} &
 \multicolumn{1}{l|}{.70} &
 \multicolumn{1}{l|}{.64} &
 \multicolumn{1}{l|}{.69} &
 \multicolumn{1}{l|}{.44} &
 \multicolumn{1}{l|}{.69} &
 \multicolumn{1}{l|}{.57} &
 \multicolumn{1}{l|}{.71} &
 \multicolumn{1}{l|}{.57} &
 \multicolumn{1}{l|}{.73} &
 \multicolumn{1}{l|}{.48} &
 \multicolumn{1}{l|}{.66} &
 \multicolumn{1}{l|}{.34} &
 \multicolumn{1}{l|}{.58} &
 \multicolumn{1}{l|}{.37} &
 \multicolumn{1}{l|}{.62} &
 \multicolumn{1}{l|}{.27} &
 \multicolumn{1}{l|}{.51} &
 \multicolumn{1}{l|}{\textit{.17}} \\ \hline
\multicolumn{1}{|l|}{tmpN2V} &
 \multicolumn{1}{l|}{.99} &
 \multicolumn{1}{l|}{.99} &
 \multicolumn{1}{l|}{.99} &
 \multicolumn{1}{l|}{.99} &
 \multicolumn{1}{l|}{.99} &
 \multicolumn{1}{l|}{.99} &
 \multicolumn{1}{l|}{.99} &
 \multicolumn{1}{l|}{.99} &
 \multicolumn{1}{l|}{.64} &
 \multicolumn{1}{l|}{.66} &
 \multicolumn{1}{l|}{.66} &
 \multicolumn{1}{l|}{.66} &
 \multicolumn{1}{l|}{.45} &
 \multicolumn{1}{l|}{.70} &
 \multicolumn{1}{l|}{.63} &
 \multicolumn{1}{l|}{.70} &
 \multicolumn{1}{l|}{.66} &
 \multicolumn{1}{l|}{.73} &
 \multicolumn{1}{l|}{.55} &
 \multicolumn{1}{l|}{.71} &
 \multicolumn{1}{l|}{.46} &
 \multicolumn{1}{l|}{.57} &
 \multicolumn{1}{l|}{.50} &
 \multicolumn{1}{l|}{.62} &
 \multicolumn{1}{l|}{.32} &
 \multicolumn{1}{l|}{.60} &
 \multicolumn{1}{l|}{\textit{.08}} \\ \hline
\multicolumn{1}{|l|}{\textit{O-A}} &
 \multicolumn{2}{c|}{\textit{.07}} &
 \multicolumn{2}{c|}{\textit{.12}} &
 \multicolumn{2}{c|}{\textit{.06}} &
 \multicolumn{2}{c|}{\textit{-.04}} &
 \multicolumn{2}{c|}{\textit{.04}} &
 \multicolumn{2}{c|}{\textit{.05}} &
 \multicolumn{2}{c|}{\textit{.14}} &
 \multicolumn{2}{c|}{\textit{.07}} &
 \multicolumn{2}{c|}{\textit{.08}} &
 \multicolumn{2}{c|}{\textit{.09}} &
 \multicolumn{2}{c|}{\textit{.11}} &
 \multicolumn{2}{c|}{\textit{.12}} &
 \multicolumn{2}{c|}{\textit{.15}} &
 \textit{} \\ \cline{1-27}
\end{tabular}
\end{table}

After obtaining the result that alignment improves the accuracy in the network inference task, we look at the relationship between the alignment errors and the amount of improvement in classification accuracy. To this end, we calculate partial Spearman's rank correlations between the absolute improvement and alignment errors (rotation and translation errors). First, we control for the translation error and look at the correlation between the improvement and rotation error. Then, we do the same by switching the control variable. The partial correlation of improvement to rotation error is $0.65$ ($p < .000001$) and to translation error is $0.42$ ($p < .000001$). This result formally shows that there is a positive relationship between the alignment errors and the amount of improvement that can be obtained by ensuring alignment.

Table \ref{tab:rw_improv_align} provides a closer look into the classification accuracy improvements on the three largest networks we use in this study. On the columns, we denote the classification task (dataset, target variable, and consecutive timesteps) and whether the model is tested on the original (O) or aligned (A) embeddings. On the rows, we list the static and dynamic embedding methods separately. We provide average improvements (O-A) over methods (the rightmost column) or over specific classification tasks (in the bottom rows after static and dynamic embedding methods). We observe that the improvement obtained for static embedding methods are larger than the improvement obtained for the dynamic embedding cases. This is in line with the earlier finding that static methods naturally produce more alignment errors hence more room for improvement. In specific cases, we observe that when the embeddings are misaligned, the classification accuracy is as low as \%1 but can jump to more than \%70 once the same embeddings are aligned.

\begin{figure}[!h]
\begin{center}
\includegraphics[width=.98\textwidth]{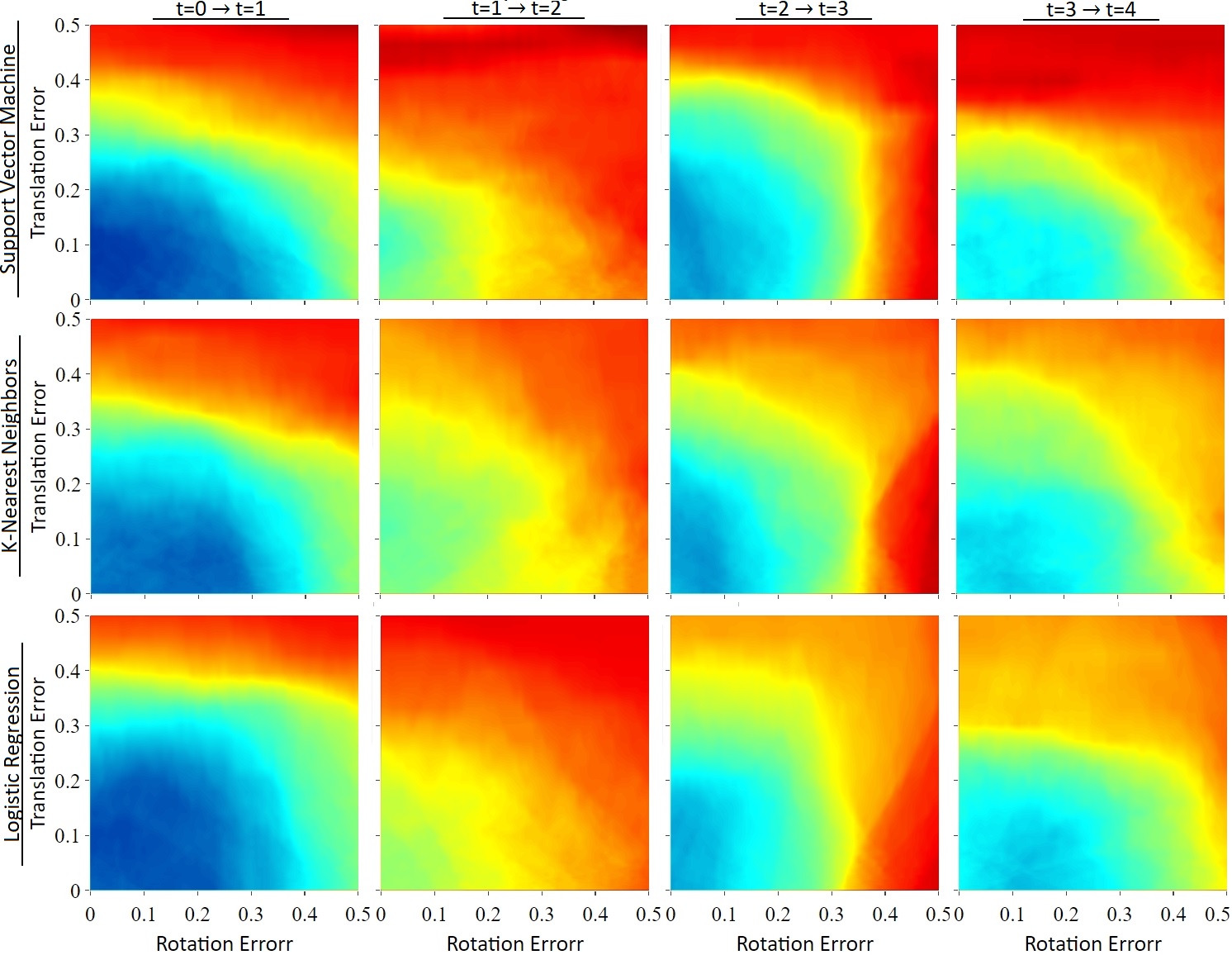}
\caption{Effect of synthetic misalignment on prediction accuracy. Each row presents a different classifier. Each column presents different consecutive time pairs for the temporal inference task. On the \textit{sp-high-school} dataset, the classifier is trained at time $t$ using the \textit{node2vec} embeddings as input features and students' class as the target variable. Accuracy is calculated on the predictions made at $t+1$ using $t+1$ embeddings.}
\label{fig:enforced_misalignment}
\end{center}
\end{figure}

Next, we take interest in the relationship between controlled forced misalignment and prediction accuracy. For the sake of simplicity, we consider only one dataset and embedding method. For each pair of consecutive timesteps of the \textit{sp-high-school} dataset, we align the corresponding \textit{node2vec} embeddings. Then, in a similar fashion to the experiments in Section \ref{syntheticNetworks}, we misalign them by applying different rotations and translations with regards to rotation and translation factors. Finally, we predict the labels of the second embeddings set using different models trained on the first set. Figure \ref{fig:enforced_misalignment} exposes the obtained results. Overall, we observe that the larger are the rotations/translations applied, the worse is getting the classification accuracy. Also, we remark that the best prediction accuracy is obtained when the rotation and translation are small. This means that, beyond helping improving performances, the way we align is ideal regarding the classification task we address for this dataset. Another interesting point lies in the similarity between the different classifier results. Thus, this results do not depend on what classification model is used. It also justifies our prior choice of employing SVM.

\begin{figure}[!h]
\begin{center}
\includegraphics[width=0.375\textwidth]{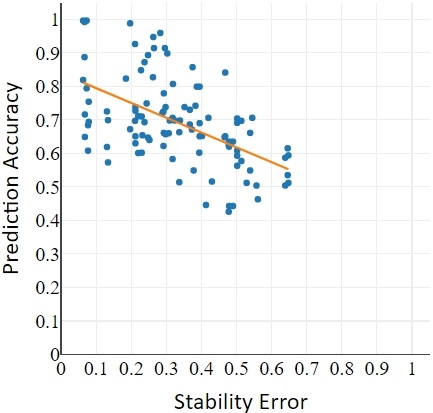}
\caption{Prediction accuracy versus stability error of the embeddings. For all the considered datasets and embedding methods, the stability measure and the prediction accuracy are reported for each pair of consecutive timestep embeddings. Globally, low stability error coincides with good prediction scores.}
\label{fig:pred_stab}
\end{center}
\end{figure}

We also look at the relationship between embedding stability and prediction accuracy in the inference task. By definition, stability error is directly related to the changes between nodes' consecutive embeddings that are not attributable to misalignment. This means that, when the stability error is high, the boundaries discriminating between the different labels of nodes would also change significantly. Consequently, the score of the prediction of nodes labels knowing their embeddings at a timestep using a model trained on the previous timestep data should be correlated to the stability error. Figure \ref{fig:pred_stab} shows the results of this experiment on datasets that have more than 50 nodes. Overall, we can see that the higher the stability error, the less accurate the prediction task is, thus confirming our previous intuition. However, we also observe that the points are somehow scattered around the linear regression line. One explanation lies in the differences between the datasets and the embedding algorithms. For example, there are cases where the classifier is performing relatively well even though the stability error is relatively high. This is possibly due to the fact that the different clusters representing the labels are separate enough to compensate for the moves of node embeddings. On the other hand, when the clusters are too close or even overlapping, it may cause misclassifications although stability error is low.

\section{Conclusion}
In this paper, we proposed a method to force alignment as well as a measuring procedure able to describe the alignment and stability of dynamic network embeddings. In order to demonstrate the validity of the developed measures, we conducted different experiments using synthetic data, proving that each one of the measures captures the type of misalignment it is designed for. Then we confronted our procedure to real-world datasets, using several static and dynamic embedding methods. Overall, we confirm that, as expected, dynamic methods produce more aligned embeddings comparing to static ones. Also, the performed experiments show that aligning embeddings generally improves prediction performances for the inference tasks that are sensitive to embeddings temporal evolution. In addition, the method we employ to align embeddings seems to be 
generally ideal
regarding the considered transformations and inference task. Finally, we demonstrate that the stability measure we propose is correlated to the accuracy of the considered inference task.

In this paper, we focused on temporal network embeddings. However, our work can be easily adapted to other types of data. It could be any sequence of matrices of the same shape, each one representing a set of vectors in an N-dimensional space, such as dynamic word embeddings or sequence of images. On a global note, as shown in this paper, aligning these data sequences can help improving their quality for different purposes. Further work tied up to aligned embeddings exploitation (dynamics analysis and prediction, visualization) is under way.

\section*{Acknowledgements}
This work is supported by the Fellowship for Young Visiting Researchers program of the Embassy of France in Turkey, IMT Atlantique, DSI Global Services, the French public agency ANRT (Association Nationale de la Recherche et de la Technologie), The Scientific and Technological Research Council of Turkey (TUBITAK) under 2211-A Program, and the Council of Higher Education (YÖK) in Turkey under 100/2000 Program.

\bibliographystyle{elsarticle-num-names}
\bibliography{sample.bib}

\end{document}